%% file: ms.tex
\documentclass[letterpaper]{article} 
\usepackage{aaai24}  
\nocopyright
\usepackage{times}  
\usepackage{helvet}  
\usepackage{courier}  
\usepackage[hyphens]{url}  
\usepackage{graphicx} 
\usepackage{natbib}  
\usepackage{caption} 
\usepackage{algorithm}
\usepackage{algorithmic}
\usepackage{newfloat}
\usepackage{listings}
\usepackage{amsmath}
\usepackage{amssymb}
\usepackage{subcaption}
\usepackage{caption}
\usepackage{booktabs}
\usepackage{multirow}
\usepackage{xcolor}
\usepackage{makecell}
\usepackage{xspace}
\usepackage{xcolor,colortbl}
\usepackage[symbol]{footmisc}
\newcommand{\etal}{\textit{et al.\xspace}}

\newcommand{\ie}{\textit{i.e.\xspace}}

\definecolor{Gray}{rgb}{0.9255,0.9255,0.9176}
\definecolor{LightCyan}{rgb}{0.7569,0.7961,0.8431}
\newcommand{\modelname}{Cas-DM\xspace}
\DeclareCaptionStyle{ruled}{labelfont=normalfont,labelsep=colon,strut=off} 
\floatstyle{ruled}
\newfloat{listing}{tb}{lst}{}
\floatname{listing}{Listing}
\title{Bring Metric Functions into Diffusion Models}

\author{
    Jie An\textsuperscript{\rm 1}\thanks{Work done during internship at Microsoft.}, Zhengyuan Yang\textsuperscript{\rm 2}, 
    Jianfeng Wang\textsuperscript{\rm 2}, 
    Linjie Li\textsuperscript{\rm 2},
    Zicheng Liu\textsuperscript{\rm 2},
    Lijuan Wang\textsuperscript{\rm 2}, 
    Jiebo Luo\textsuperscript{\rm 1}
}
\affiliations{
    \textsuperscript{\rm 1}University of Rochester,
    \textsuperscript{\rm 2}Microsoft Azure AI

    \tt\small{\{jan6,jluo\}@cs.rochester.edu},\\ 
    \tt\small\{zhengyang,jianfw,lindsey.li,zliu,lijuanw\}@microsoft.com
}

\usepackage{bibentry}

\begin{document}

\maketitle

\input{sections/abstract_v2.tex}

\input{sections/introduction_v2.tex}

\input{sections/related_work.tex}
\input{sections/background}

\input{sections/method.tex}

\input{sections/experiment.tex}

\input{sections/conclusion.tex}

\bibliography{ms}

\end{document}

%% file: sections/abstract_v2.tex
\begin{abstract}
    We introduce a Cascaded Diffusion Model (\textbf{\modelname}) that improves a Denoising Diffusion Probabilistic Model (DDPM) by effectively incorporating additional metric functions in training. 
    Metric functions such as the LPIPS loss have been proven highly effective in consistency models derived from the score matching.
    However, for the diffusion counterparts, the methodology and efficacy of adding extra metric functions remain unclear.
    One major challenge is the mismatch between the noise predicted by a DDPM at each step and the desired clean image that the metric function works well on.
    To address this problem, we propose \modelname, a network architecture that cascades two network modules to effectively apply metric functions to the diffusion model training.
    The first module, similar to a standard DDPM, learns to predict the added noise and is  unaffected by the metric function. The second cascaded module learns to predict the clean image, thereby facilitating the metric function computation.
    Experiment results show that the proposed diffusion model backbone enables the effective use of the LPIPS loss, leading to state-of-the-art image quality (FID, sFID, IS) on various established benchmarks.
\end{abstract}

%% file: sections/introduction_v2.tex
\begin{figure}[t!]
     \centering
     \begin{subfigure}[b]{\linewidth}
         \centering
         \includegraphics[width=.8\textwidth]{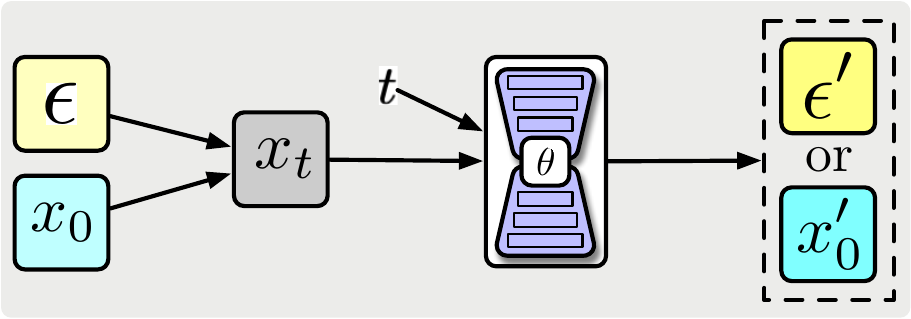}
         \caption{DDPM~\cite{ho2020denoising}}
         \label{fig:intro_ddpm}
     \end{subfigure}
     \hfill
     \begin{subfigure}[b]{\linewidth}
         \centering
         \includegraphics[width=.8\textwidth]{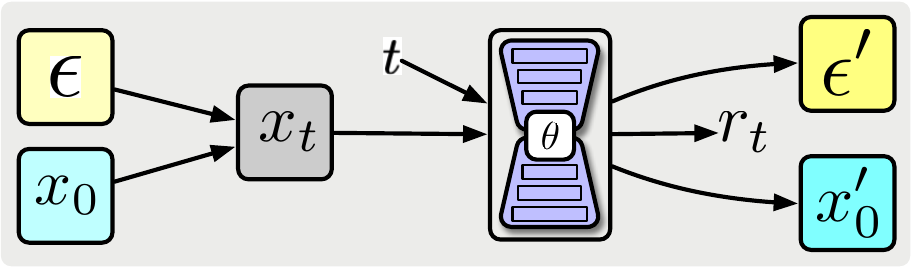}
         \caption{Dual Diffusion Model~\cite{benny2022dynamic}}
         \label{fig:intro_dual}
     \end{subfigure}
     \hfill
     \begin{subfigure}[b]{\linewidth}
         \centering
         \includegraphics[width=.8\textwidth]{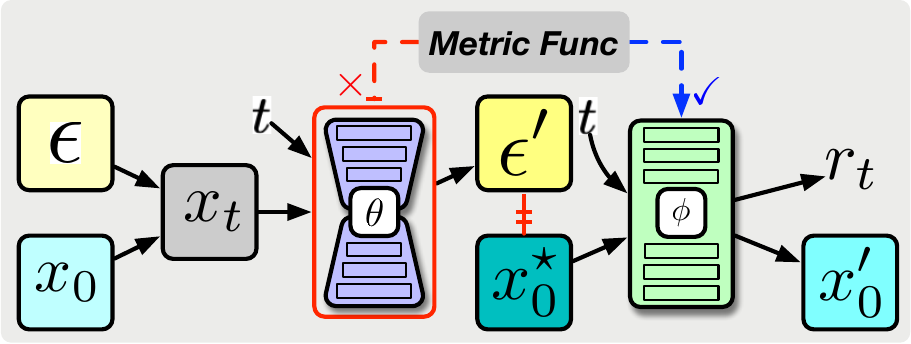}
         \caption{Ours}
         \label{fig:intro_ours}
     \end{subfigure}
        \caption{
        We introduce a cascaded diffusion model that can effectively incorporate metric functions in diffusion training.
        \textbf{(a)} DDPM outputs either $\epsilon^\prime$ or $x_0^\prime$ and uses the corresponding loss in training. \textbf{(b)} Dual Diffusion Model outputs both $\epsilon^\prime$ and $x_0^\prime$ simultaneously with a single network $\theta$, where applying metric functions on $x_0^\prime$ will inevitably influence the prediction of $\epsilon^\prime$. \textbf{(c)} Our \modelname cascades the main module $\theta$ with an extra network $\phi$, where $\theta$ is frozen for the $x_0^\prime$-related losses and metric functions.}
        \label{fig:intro_training}
\end{figure}

\section{Introduction}
\label{sec:introduction}
The Denoising Diffusion Probabilistic Model (DDPM)~\cite{ho2020denoising} has emerged as a leading method in visual content generation, positioned among other approaches such as Generative Adversarial Networks (GAN)~\cite{goodfellow2014generative}, Variational Auto-Encoders (VAE)~\cite{kingma2013auto}, auto-regressive models~\cite{esser2021taming}, and normalization flows~\cite{kingma2018glow}. 
DDPM is a score-based model that adopts an iterative Markov chain in generating images, where the transition of the chain is the reverse diffusion process to gradually denoise images. 

Recently, \citet{song2023consistency} propose a novel score-based generative model called consistency model. One key observation is that using metric functions such as the Learned Perceptual Image Patch Similarity (LPIPS) loss~\cite{zhang2018perceptual} in training can significantly improve the quality of generated images. 
The LPIPS loss, with its VGG backbone~\cite{simonyan2014very} trained on the ImageNet dataset for classification, allows the model to capture more accurate and diverse semantic features, which may be hard to learn through generative model training alone.
However, it remains unclear whether adding additional metric functions could yield similar improvements in diffusion models. In this study, taking LPIPS loss as a prototype, we explore how to effectively incorporate metric functions into diffusion models.
The primary challenge lies in the mismatch between the multi-step denoising process that generates noise predictions, and the single-step metric function computation that requires a clean image.


We next zoom in on the DDPM process to better illustrate this mismatch challenge. As shown in Figure~\ref{fig:intro_training},
DDPM adopts a diffusion process to gradually add noise to a clean image $x_0$, producing a series of noisy images $x_i, i\in 1,...,T$. Then the model is trained to perform a reverse denoising process by predicting a less noisy image $x_{t-1}$ from $x_t$. 
Instead of directly predicting $x_{t-1}$, DDPM gives two ways to obtain $x_{t-1}$: predicting either the clean image $x_0$ or the added Gaussian noise $\epsilon_t$.
The training objective of the DDPM is the mean squared error (MSE)
between the predicted and ground truth $x_0$ or $\epsilon_t$, where a few papers~\cite{nichol2021improved,benny2022dynamic} found the latter (\ie, the $\epsilon$ mode) to be empirically better than predicting $x_0$ (\ie, the $x_0$ mode).
The two modes in DDPM provide us with two initial options for bringing metric functions. Applying metric functions directly to predicted noise $\epsilon$ is unreasonable because the networks for metric functions are trained on RGB images and produce meaningless signals when applied to noise $\epsilon$. Despite the promising improvements, this $x_0$-mode model with metric functions still suffers from the low performance from the $x_0$-mode baseline, when compared with the $\epsilon$-mode.
This naturally motivates the question: {can we merge the two modes and further improve the $\epsilon$-mode performance with the metric function?} To achieve this, we need a diffusion model that can generate $x-0$ while maintaining the $\epsilon$-mode performance. The goal is made possible with the Dynamic Dual Diffusion Model~\cite{benny2022dynamic}, where authors expand the output channel of the DDPM's network $\theta$ to let it predict $x_0$, $\epsilon$, and a dynamic mixing weight, simultaneously. The experiments show that Dual Diffusion Model outperforms both the $x_0$- and $\epsilon$-modes of DDPM. However, naively adding the metric function to its $x_0$ head will not work. This is because the additional metric functions on the predicted $x_0$ updates the shared backbone, which disturbs the $\epsilon$ prediction and leads to degraded performance.

To this end, we propose a new Cascaded Diffusion Model (\textbf{\modelname}), which allows the application of metric functions to DDPM by addressing the above-mentioned issues. We cascade two network modules, where the first model $\theta$ takes the noisy image $x_t$ and predicts the added noise.
We then derive an initial estimation of $x_0$
based on $x_t$ and $\epsilon_{\theta}$ following equations of the diffusion process. Next, the second model $\phi$ takes the initial $x_0$ prediction 
and the time step $t$ and output the refined prediction of $x_0$ as well as the dynamic weight to mix $x_0$ and $\epsilon$ predictions in diffusion model sampling.
In training, we apply the metric function to the predicted $x_0$ of $\phi$, which is used to update the parameters of $\phi$ and 
stop the gradient for $\theta$.
This ensures the $\epsilon$ branch to be intact while the $x_0$ branch is enhanced by the additional metric function. 

Experimental results on CIFAR10~\cite{krizhevsky2009learning}, CelebAHQ~\cite{karras2017progressive}, LSUN-Church/Bedroom~\cite{yu2015lsun}, and ImageNet~\cite{deng2009imagenet} show that applying the LPIPS loss on \modelname~can effectively improve its performance, leading to the state-of-the-art image quality (measured by FID~\cite{heusel2017gans}, sFID~\cite{nash2021generating}, and IS~\cite{salimans2016improved}) on most datasets. This work demonstrates that with a careful architecture design, metric functions such as the LPIPS loss can also be used to improve the performance of diffusion models.

Our contributions are three-fold:
\begin{itemize}
    \item We explore the methodology and efficacy of introducing extra metric functions into DDPM, resulting in a framework that can effectively incorporate metric functions during diffusion training.
    \item We introduce \modelname~that addresses the main challenge in adding metric functions to DDPM by jointly predicting the added noise and the original clean image in each diffusion training and denoising step.
    \item Experiment results show that \modelname~with the LPIPS loss consistently outperforms the state of the art across various datasets with different sampling steps.
\end{itemize}


%% file: sections/related_work.tex
\begin{figure*}[t]
    \centering
    \includegraphics[width=.9\textwidth]{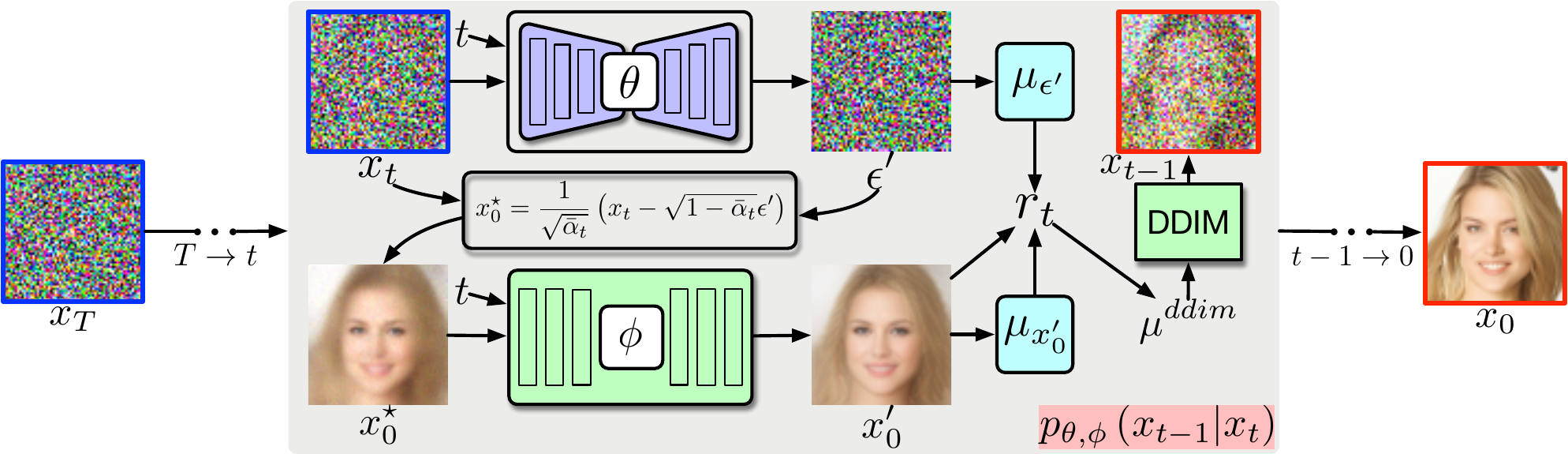}
    \caption{Framework of the proposed \modelname. For each time step $t$ from $T$ to $1$, $\theta$
    takes $x_t$ and $t$ as the inputs and estimates the added noise $\epsilon^\prime$, which is then converted into an estimation of the clean image $x_0^\star$. Next, $\phi$ outputs the $x_0^\prime$ and $r_t$ based on $x_0^\star$ and $t$, where the former is the final clean image estimation. $r_t$ is then used to mix the $\mu$ estimations from $x_0^\prime$ and $\epsilon^\prime$. \modelname uses DDIM to run one backward step based on $\mu^{ddim}$, getting $x_{t-1}$. \modelname runs the above process for $T-1$ rounds and gradually generates a clean image starting from a noise sample.
    }
    \label{fig:framework}
\end{figure*}

\section{Related Work}
\label{sec:related_work}

\noindent\textbf{Denoising Diffusion Probabilistic Models.}
Starting from DDPM introduced by Ho~\etal, diffusion models~\cite{ho2020denoising,dhariwal2021diffusion,nichol2021improved,rombach2022high} have outperformed GANs~\cite{goodfellow2014generative,karras2017progressive,mao2017least,brock2018large,wu2019logan,karras2019style,karras2020analyzing}, Variational Auto-Encoders (VAE)~\cite{kingma2013auto,van2017neural,vahdat2020nvae}, auto-regressive models~\cite{van2016pixel,van2016conditional,salimans2017pixelcnn++,chen2018pixelsnail,razavi2019generating,esser2021taming}, and normalization flows~\cite{dinh2014nice,dinh2016realnvp,kingma2018glow,ho2019flow++} in terms of image quality while having a pretty stable training process. The diffusion model is in line with the score-based~\cite{song2019generative,song2020improved} and Markov-chains-based~\cite{bengio2014deep,salimans2015markov} generative models, where the diffusion process can also be theoretically modeled by the discretization of a continuous SDE~\cite{song2020score}.
Diffusion models have been used to generate multimedia content such as audio~\cite{oord2016wavenet}, image~\cite{brock2018large,saharia2022photorealistic,ramesh2022hierarchical}, and video~\cite{singer2022make,zhou2022magicvideo,ho2022imagen,an2023latent,blattmann2023align}. The open-sourced latent diffusion model~\cite{rombach2022high} sparks numerous image generation models based on conditions such as text~\cite{saharia2022photorealistic,ramesh2022hierarchical,yang2023reco}, sketch/segmentation maps~\cite{rombach2022high,fan2023frido}, and images in distinct domains~\cite{saharia2022palette}.

\noindent\textbf{Improving Diffusion Models.} The success of the diffusion model has drawn increasing interest in improving its algorithmic design. \citet{nichol2021improved} improve the log-likelihood estimation and the generation quality of the DDPM by introducing a cosine-based noise schedule and letting the model learn variances of the reverse diffusion process in addition to the mean value in training. \citet{rombach2022high} introduce the latent diffusion model (LDM), which deploys the diffusion model on the latent space of an auto-encoder to reduce the computation cost. \citet{song2020denoising} improve the sampling speed of the diffusion model by proposing an implicit diffusion model called DDIM. In terms of the architecture design, Benny and Wolf propose Dual Diffusion Model, which learns to predict $\epsilon$ and $x_0$ simultaneously in training, leading to improved generation quality. For the training approach, \citet{jolicoeur2020adversarial} explore adopting the adversarial loss as an extra loss to improve the prediction of $x_0$. This work studies an orthogonal improvement aspect of diffusion models -- How to use additional metric functions to improve the generation performance. We draw the inspiration from~\citet{jolicoeur2020adversarial} and~\citet{benny2022dynamic}. While \citet{jolicoeur2020adversarial} found that the adversarial objective based on a learnable discriminator is unnecessary for powerful generative models, we found that metric functions based on a fixed pre-trained network can achieve improved performance with a proper network architecture and training approach. The proposed diffusion model backbone shares the same idea of the dual output as~\citet{benny2022dynamic} but has different architectures and training strategies.

%% file: sections/background.tex
\section{Preliminary}
\label{sec:background}
This section introduces the forward/backward diffusion processes and the training losses of DDPM, which will be used to derive the proposed \modelname later.
The theory of diffusion models consists of a forward and a backward process. Given a clean image $x_0$ and a constant $T$ to denote the maximum steps, the forward process gradually adds randomly sampled noise from a pre-defined distribution to $x_0$, leading to a sequence of images $x_t$ for time steps $t\in\left[1,...,T\right]$, where $x_t$ is derived by adding noise to $x_{t-1}$. DDPM~\cite{ho2020denoising} uses the Gaussian noise, resulting in the following transition equation,
\begin{equation}
    q\left(x_t | x_{t-1}\right) := \mathcal{N}\left(x_t;\sqrt{1-\beta_t}x_{t-1},\beta_t\mathbf{I}\right),
    \label{eq:1}
\end{equation}
where $\beta_t\in\left(0,1\right]$ are pre-defined constants. Eq.~\ref{eq:1} can derive a direct transition from $x_0$ to $x_t$,
\begin{equation}
    q\left(x_t | x_0\right) := \mathcal{N}\left(x_t;\sqrt{\bar{\alpha}_t}x_0,\left(1 - \bar{\alpha}_t\right)\mathbf{I}\right),
    \label{eq:2}
\end{equation}
where $\alpha_t:=1-\beta_t$ and $\bar{\alpha}_t := \prod_{i=1}^{t}\alpha_i$. Via Eq.~\ref{eq:2}, for any $t\in\left[1, T\right]$, one can easily get $x_t$ given $x_0$ and a noise sample $\epsilon \sim \mathcal{N}\left(\mathbf{0}, \mathbf{1}\right)$,
\begin{equation}
    x_t = \sqrt{\bar{\alpha}_t}x_0 + \sqrt{1 - \bar{\alpha}_t}\epsilon.
    \label{eq:3}
\end{equation}
Given a fixed $x_t$, Eq.~\ref{eq:3} bridges $x_0$ and $\epsilon$.

The backward process gradually recovers $x_0$ from the noisy image $x_T\in\mathcal{N}\left(x_T; \mathbf{0}, \mathbf{I}\right)$, where for each $t$, the transition from $x_{t}$ to $x_{t - 1}$ is 
$p\left(x_{t-1} | x_t\right)$, 
which is the ultimate target to learn of the diffusion model. 
The backward transition 
$p$
is then approximated by
\begin{equation}
    p_\theta \left(x_{t-1} | x_t\right) := \mathcal{N}\left(x_{t-1};\mu_\theta \left(x_t, t\right), \Sigma_\theta \left(x_t, t\right) \right).
\end{equation}
The training objective is to maximize the 
variational lower bound (VLB) of the data likelihood.
DDPM simplifies the training process to be first uniformly sample a $t$ from $\left[1, T\right]$ and then compute,
\begin{equation}
    L_t := D_{\mathrm{KL}}\left(q\left(x_{t-1} | x_t, x_0\right) \| p_\theta\left(x_{t-1} | x_t\right) \right),
\end{equation}
which is further simplified to be 
\begin{equation}
    L_t:=\frac{1}{2 \beta_t^2}\left\|\tilde{\mu}_t\left(x_t, x_0, t\right)-\mu_\theta\left(x_t, t\right)\right\|^2,
\end{equation}
where
\begin{equation}
\tilde{\mu}_t\left(x_t, x_0, t\right):= \frac{\sqrt{\bar{\alpha}_{t-1}} \beta_t}{1-\bar{\alpha}_t} x_0+\frac{\sqrt{\alpha_t}\left(1-\bar{\alpha}_{t-1}\right)}{1-\bar{\alpha}_t} x_t.
\label{eq:4}
\end{equation}
$\tilde{\mu}_t\left(x_t, x_0, t\right)$ and $\mu_\theta\left(x_t, t\right)$ are the mean values of $q\left(x_{t-1} | x_t, x_0\right)$ and $p_\theta\left(x_{t-1} | x_t\right)$, respectively.
DDPM parameterizes $\mu_\theta$ with a neural network $\theta$ that either predicts $x_0$ or $\epsilon$, where two types of network outputs are denoted as $x_0^\prime$ and $\epsilon^\prime$, respectively. We can obtain $\mu_\theta$ via Eq.~\ref{eq:4} with $x_0^\prime$. 
If the network predicts
$\epsilon^\prime$, we first get an indirect $x_0$ prediction from $\epsilon^\prime$ via Eq.~\ref{eq:3}. Then we can compute $\mu_\theta$ via Eq.~\ref{eq:4} with the indirect $x_0$ prediction. Ho~\etal ~empirically demonstrate that $\epsilon^\prime$ usually yields better image quality, i.e., lower FID score~\cite{heusel2017gans} than $x_0^\prime$. One may refer to ~\cite{benny2022dynamic} and~\cite{ho2020denoising} for more detailed mathematical derivation. 

%% file: sections/method.tex
\section{Method}
\label{sec:method}
This section introduces the network architecture of our \modelname~as well as its training and sampling processes.
\subsection{Cascaded Diffusion Model}
As shown in Fig.~\ref{fig:framework}, the backbone of \modelname~consists of two cascaded networks, denoted as $\theta$ and $\phi$. $\theta$ is used to predict the added noise $\epsilon$, and $\phi$ is to predict the clean image $x_0$. The architectures of both $\theta$ follow the improved diffusion~\cite{nichol2021improved} while $\phi$ is a network whose input and output tensors have the same shape. We use the model output from both $\theta$ and $\phi$ to obtain an estimation of $\mu_{\theta,\phi}\left(x_t, t\right)$, detailed as follows. 

In training, $\theta$ takes the noisy image $x_t$ and the uniformly sampled time step $t$ as the input and predict the added noise $\epsilon^\prime$,
$\theta$ is equivalent to a vanilla DDPM predicting $\epsilon^\prime$.
Based on Eq.~\ref{eq:3}, $\epsilon^\prime$ can lead to an indirect estimation of $x_0$ as follows,
\begin{equation}
    x_0^\star = \frac{1}{\sqrt{\bar{\alpha}_t}} \left(x_t - \sqrt{1 - \bar{\alpha}_t} \epsilon^\prime \right).
    \label{eq:5}
\end{equation}
Next, based on $\epsilon^\prime$, we obtain an estimation of $\mu_\theta\left(x_t, t\right)$, which is denoted as $\mu_{\epsilon^\prime}\left(x_t, t\right)$,
\begin{equation}
    \mu_{\epsilon^\prime}\left(x_t, t\right) := \frac{1}{\sqrt{\alpha_t}} x_t-\frac{1-\alpha_t}{\sqrt{1-\bar{\alpha}_t} \sqrt{\alpha_t}} \epsilon^\prime.
    \label{eq:6}
\end{equation}
Eq.~\ref{eq:6} is derived by replacing $x_0$ in Eq.~\ref{eq:4} with $x_0^\star$ in Eq.~\ref{eq:5}.

$\phi$ takes $x_0^\star$ and $t$ as the input and output $x_0^\prime$ as well as a dynamic value $r_t$, which is used to balance the strength of $\theta$ and $\phi$ in computing $\mu_{\theta,\phi}\left(x_t, t\right)$ later. The application of $r_t$ is directly inspired by dual diffusion~\cite{benny2022dynamic}, where their experiments show that $r_t$ can better balance the effects of two types of predictions and lead to improved performance. We follow this setting and obtain $r_t$ by adding an extra channel to the output layer of $\phi$. The output of $\phi$ is the concatenation of $x_0^\prime\in\mathbb{R}^{H,W,C}$ and $r_t\in\mathbb{R}^{H,W,1}$ along the channel dimension.
We obtain the estimation of $\mu_\theta\left(x_t, t\right)$ based on $x_0^\prime$ as 
\begin{equation}
\mu_{x_0^\prime}\left(x_0^\star, t\right):= \frac{\sqrt{\bar{\alpha}_{t-1}} \beta_t}{1-\bar{\alpha}_t} x_0^\prime+\frac{\sqrt{\alpha_t}\left(1-\bar{\alpha}_{t-1}\right)}{1-\bar{\alpha}_t} x_t.
\end{equation}
$\phi$ is to improve the accuracy of the $x_0$ prediction on top of $\theta$'s output.
The final estimation of $\mu_{\theta,\phi}\left(x_t, t\right)$ is
\begin{equation}
    \mu_{\theta,\phi}\left(x_t, t\right) = r_t \cdot \mu_{x_0^\prime}\left(x_t, t\right) + \left( 1 - r_t \right) \cdot \mu_{\epsilon^\prime}\left(x_t, t\right)
    \label{eq:7}
\end{equation}

Compared with dual diffusion~\cite{benny2022dynamic}, \modelname~allows the dedicated metric functions to be applied on the $x_0$ branch without influencing the $\epsilon$ branch because we could stop the gradients of metric functions on $\theta$. More details will be in the next part.

\begin{figure}[t]
    \centering
    \includegraphics[width=\linewidth]{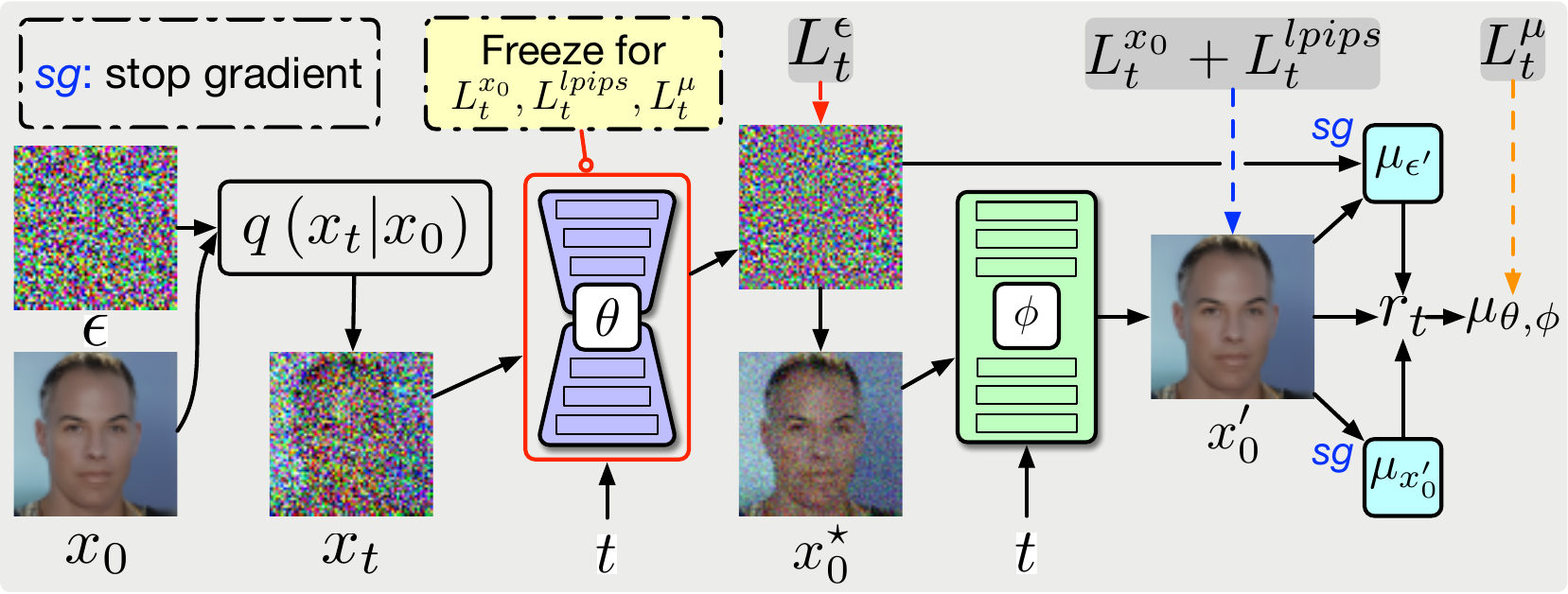}
    \caption{Training process of \modelname. $\theta$ learns to estimate the added noise $\epsilon$ while $\phi$ is trained to predict the clean image $x_0$. We apply $L_t^\epsilon$ on $\theta$ and all the gradients of other losses are blocked for it. For $\phi$, we use $L_t^{x_0}$, $L_t^{lpips}$, and $L_t^{\mu}$ losses, where the first two is to enforce $\phi$ to recover the clean image from $x_0^\star$, assisted by the the LPIPS loss. $L_t^{\mu}$ is to train the dynamic mixing weight and the gradient is stopped before $\mu_{\epsilon^\prime}$ and $\mu_{x_0^\prime}$. Best viewed on screen by zoom-in. }
    \label{fig:method_training}
\end{figure}

\subsection{Training and Sampling}
As shown in Fig.~\ref{fig:method_training}, we train \modelname~following the approach of dual diffusion~\cite{benny2022dynamic} with the following loss terms,
\begin{equation}
\begin{gathered}
L_t^\epsilon=\left\|\epsilon-\epsilon^\prime\right\|^2, \\
L_t^{x_0}=\left\|x_0-x_0^\prime\right\|^2, \\
L_t^\mu=\left\|\tilde{\mu}_t-\left(r_t\left[\mu_{x_0^\prime}\right]_{\mathrm{sg}}+\left(1-r_t\right)\left[\mu_{\epsilon^\prime}\right]_{\mathrm{sg}}\right)\right\|^2. \\
\end{gathered}
\end{equation}
The input value of $\mu_t$, $\mu_{x_0^\prime}$, and $\mu_{\epsilon^\prime}$ are omitted for simplicity. $\left[\cdot\right]_{\mathrm{sg}}$ denotes stop gradients.
We use the LPIPS loss~\cite{johnson2016perceptual} from the \texttt{piq} repository\footnote{https://github.com/photosynthesis-team/piq} in training to demonstrate that extra metric functions can be applied to \modelname~for further improvements,
\begin{equation}
    L_t^{lpips}=\mathrm{LPIPS}\left(\mathcal{T}\left(x_0\right), \mathcal{T}\left(x_0^\prime\right)\right).
\end{equation}
Here $\mathcal{T}$ denotes an image transformation module, which first interpolates an image to the size of $224\times 224$ with the bilinear interpolation, then linearly normalize its value to the range of $\left[0, 1\right]$.
In back-propagating, we disconnect $\theta$ and $\phi$ by detaching the whole $\theta$ from the computing graph of $\phi$, leading to separate loss functions for $\theta$ and $\phi$:
\begin{equation}
    L_t^{\theta}=\lambda^\epsilon L_t^\epsilon,
\end{equation}
\begin{equation}
    L_t^{\phi}=\lambda^{x_0} L_t^{x_0}+\lambda^\mu L_t^\mu +\lambda^{lpips} L_t^{lpips}. 
\end{equation}
Since the LPIPS loss only works well on real images, we use $L_t^{\theta}$ to let $\theta$ learn to predict $\epsilon$ without the disturbance of the gradient from $\phi$ and the metric functions, leading to a stable $\mu_\theta$ estimation, $\mu_{\epsilon^\prime}$, as the basis. On top of it, $\phi$ learns to predict $x_0$, resulting in another estimation $\mu_{\epsilon^\prime}$. The LPIPS loss can improve the accuracy of $\mu_{\epsilon^\prime}$, leading to an overall better $\mu_\theta$ estimation through Eq.~\ref{eq:7}.

We use DDIM~\cite{song2020denoising} for sampling. Following dual diffusion~\cite{benny2022dynamic}, we obtain the $\mu$ estimation via the DDIM's $\mu$ computing equation from $\epsilon^\prime$ and $x_0^\prime$, respectively,
\begin{equation}
    \mu_{x_0^\prime}^{ddim}=\sqrt{\bar{\alpha}_{t-1}} x_0^\prime + \sqrt{1-\bar{\alpha}_{t-1}-\sigma_t^2} \cdot \frac{x_t-\sqrt{\bar{\alpha}_t} x_0^\prime}{\sqrt{1-\bar{\alpha}_t}},
\end{equation}
\begin{equation}
    \mu_{\epsilon^\prime}^{ddim}=\frac{x_t-\sqrt{1-\bar{\alpha}_t} \epsilon^\prime}{\sqrt{\alpha_t}}+\sqrt{1-\bar{\alpha}_{t-1}-\sigma_t^2} \cdot \epsilon^\prime.
\end{equation}
Then the final estimation of the $\mu^{ddim}$ for DDIM sampling is the interpolation of $\mu_{x_0^\prime}^{ddim}$ and $\mu_{\epsilon^\prime}^{ddim}$ based on $r_t$.

%% file: sections/experiment.tex
\begin{table*}[t]

    \scriptsize
    \begin{subtable}[t]{.5\linewidth}
    \setlength\tabcolsep{4pt}
    \begin{center}
    \begin{tabular}[t]{l|ccc}
    \toprule
    \rowcolor{Gray}\textbf{Model}\hspace{50mm} & \textbf{FID}$\downarrow$ & \textbf{sFID}$\downarrow$ & \textbf{IS}$\uparrow$ \\
    \\
    \multicolumn{4}{l}{\bf{CIFAR10} 32$\times$32} \\
    \toprule
    $^\star$Gated PixelCNN~\cite{van2016conditional}         & 65.93 & - & 4.60 \\
    $^\star$EBM~\cite{song2021train}                    & 38.20 & - & 6.78 \\
    $^\star$NCSNv2~\cite{song2020improved}                 & 31.75 & - & - \\
    $^\star$SNGAN-DDLS~\cite{che2020your}             & 15.42 & - & 9.09 \\
    $^\star$StyleGAN2 + ADA (v1)~\cite{karras2020analyzing}   & 3.26  & - & 9.74 \\
    $^\star$DDPM~\cite{ho2020denoising}                 & 32.65 & - & - \\
    $^\star$DDIM~\cite{song2020denoising}                 & 5.57 & - & - \\
    $^\star$Improved DDPM~\cite{nichol2021improved}        & 4.58 & - & - \\
    $^\star$Improved DDIM~\cite{nichol2021improved}        & 6.29 & - & - \\
    $^\star$Dual Diffusion~\cite{benny2022dynamic}       & 5.10 & - & - \\
    $^\star$Consistency Model (CD)~\cite{benny2022dynamic}       & 2.93 & - & 9.75 \\
    $^\star$Consistency Model (CT)~\cite{benny2022dynamic}       & 5.83 & - & 8.85 \\
    
    \midrule
    \rowcolor{Gray}DDPM ($\epsilon$ mode)        & 6.79 & 4.97 & 8.76 \\
    \rowcolor{Gray}DDPM ($x_0$ mode)             & 17.78 & 6.69 & 7.74 \\
    \rowcolor{Gray}DDPM ($x_0$ + LPIPS)          & 9.34 & 6.94 & \underline{8.83} \\
    \rowcolor{Gray}Dual Diffusion                & 6.52 & \textbf{4.60} & 8.72 \\
    \rowcolor{Gray}Dual Diffusion + LPIPS        & \textbf{5.65} & 4.89 & 8.76 \\
    \rowcolor{LightCyan}\textbf{\modelname}           & 6.80 & 5.03 & \textbf{8.88} \\
    \rowcolor{LightCyan}\textbf{\modelname + LPIPS}   & \underline{6.40} & \underline{4.87} & 8.69 \\
    \midrule
    \\
    \multicolumn{4}{l}{\bf{LSUN Bedroom} 64$\times$64} \\
    \toprule
    
    \rowcolor{Gray}DDPM ($\epsilon$ mode)        & 5.51 & \underline{27.61} & \textbf{2.13} \\
    \rowcolor{Gray}DDPM ($x_0$ mode)             & 10.28 & 32.13 & 1.85 \\
    \rowcolor{Gray}DDPM ($x_0$ + LPIPS)          & 13.14 & 32.93 & 1.70 \\
    \rowcolor{Gray}Dual Diffusion                & 5.49 & 27.71 & 2.00 \\
    \rowcolor{Gray}Dual Diffusion + LPIPS        & 7.72 & 29.08 & 1.84 \\
    \rowcolor{LightCyan}\textbf{\modelname}           & \underline{5.29} & 27.80 & \underline{2.06} \\
    \rowcolor{LightCyan}\textbf{\modelname + LPIPS}   & \textbf{5.17} & \textbf{27.45} & 2.01 \\
    \bottomrule
    
    \end{tabular}
    \end{center}
    \end{subtable}%
    \begin{subtable}[t]{.5\linewidth}
    \setlength\tabcolsep{4pt}
    \begin{center}
    \begin{tabular}[t]{l|ccc}
    \toprule
    \rowcolor{Gray}\textbf{Model} & \textbf{FID}$\downarrow$ & \textbf{sFID}$\downarrow$ & \textbf{IS}$\uparrow$ \\
    \\
    \multicolumn{4}{l}{\bf{CelebAHQ} 64$\times$64} \\
    \toprule

    $^\star$DDPM~\cite{ho2020denoising}                 & 43.90 & - & - \\
    $^\star$DDIM~\cite{song2020improved}                 & 6.15 & - & - \\
    $^\star$Dual Diffusion~\cite{benny2022dynamic}       & 4.07 & - & - \\

    \midrule
    
    \rowcolor{Gray}DDPM ($\epsilon$ mode)        & 6.34 & 17.16 & \underline{2.43} \\
    \rowcolor{Gray}DDPM ($x_0$ mode)             & 8.82 & 19.11 & 2.30 \\
    \rowcolor{Gray}DDPM ($x_0$ + LPIPS)          & 10.54 & 21.00 & 2.18 \\
    \rowcolor{Gray}Dual Diffusion                & 5.47 & 15.26 & 2.35 \\
    \rowcolor{Gray}Dual Diffusion + LPIPS        & 6.86 & 16.06 & 2.25 \\
    \rowcolor{LightCyan}\textbf{\modelname}           & \underline{5.33} & \underline{14.87} & \textbf{2.47} \\
    \rowcolor{LightCyan}\textbf{\modelname + LPIPS}   & \textbf{4.95} &\textbf{14.71} & 2.37 \\
    \midrule

    \vspace{-1.7mm}
    \\
    \multicolumn{4}{l}{\bf{ImageNet} 64$\times$64} \\
    \toprule

    \multicolumn{4}{l}{\bf{\textit{with guidance}}} \\
    \midrule
    
    $^\star$BigGAN-deep~\cite{brock2018large}     & 4.06 & 3.96 & - \\
    $^\star$Improved DDPM~\cite{nichol2021improved}   & 2.92 & 3.79 & - \\
    $^\star$ADM~\cite{dhariwal2021diffusion}             & 2.61 & 3.77 & - \\
    $^\star$ADM (dropout)~\cite{dhariwal2021diffusion}   & 2.07 & 4.29 & - \\

    \midrule
    \multicolumn{4}{l}{\bf{\textit{without guidance}}} \\
    \midrule
    $^\star$Consistency Model (CD)~\cite{benny2022dynamic}       & 4.70 & - & - \\
    $^\star$Consistency Model (CT)~\cite{benny2022dynamic}       & 11.10 & - & - \\
    
    \rowcolor{Gray}DDPM ($\epsilon$ mode)        & \underline{27.96} & 18.73 & \textbf{13.34} \\
    \rowcolor{Gray}DDPM ($x_0$ mode)             & 65.09 & 23.86 & 8.87 \\
    \rowcolor{Gray}DDPM ($x_0$ + LPIPS)          & 41.41 & 28.20 & 11.30 \\
    \rowcolor{Gray}Dual Diffusion                & 38.65 & \underline{18.38} & 11.11 \\
    \rowcolor{Gray}Dual Diffusion + LPIPS        & 31.84 & 21.88 & 12.44 \\
    \rowcolor{LightCyan}\textbf{\modelname}           & 28.34 & 18.46 & \underline{13.05} \\
    \rowcolor{LightCyan}\textbf{\modelname + LPIPS}   & \textbf{27.54} & \textbf{18.06} & 12.94 \\
    \bottomrule
    
    \end{tabular}
    \end{center}
    \end{subtable}
    \caption{Performance comparison of \modelname and the baseline models on CIFAR10, CelebAHQ, LSUN Bedroom, and ImageNet. The best and second best results are marked with \textbf{bold} and \underline{underline}, respectively. Models marked with $\star$ are borrowed from~\citet{ho2020denoising}, \citet{benny2022dynamic}, \citet{dhariwal2021diffusion}, and \citet{song2023consistency}, which are \textbf{for reference and not directly comparable} with other models due to the different diffusion model implementation, training and sampling settings, dataset preparation, and FID evaluation settings.}
    \label{tab:sota}
\end{table*}


\begin{table}[t]
    \centering
    \setlength\tabcolsep{4pt}
    \resizebox{1\linewidth}{!}{
    \begin{tabular}{l|cccc}
    \toprule
    \textbf{Model} & \textbf{CIFAR10} & \textbf{CelebAHQ} & \textbf{\thead{LSUN \\ Bedroom}} & \textbf{ImageNet} \\
    \midrule
    DDPM ($x_0$ mode)      & 17.78 & 8.82 & 10.28 & 65.09 \\
    \hspace{5mm} + LPIPS   & -8.44 & +1.72 & +2.86 & -24.68 \\
    \midrule
    Dual Diffusion         & 6.52 & 5.47 & 5.49 & 38.65 \\
    \hspace{5mm} + LPIPS   & -0.87 & +1.39 & +2.23 & -6.81 \\
    \midrule
    \textbf{\modelname}    & 6.80 & 5.33 & 5.29 & 28.34 \\
    \hspace{5mm} + LPIPS   & -0.40 & -0.38 & -0.12 & -0.8 \\
    \bottomrule
    \end{tabular}
    }
    \caption{FID variation comparison after applying the LPIPS loss. 
    The FID values of DDPM and Dual Diffusion Model fluctuate after applying the LPIPS loss. \modelname achieves consistent improvement across all the compared datasets.}
    \label{tab:lpips}
\end{table}


\section{Experiments}
\label{sec:experiment}

\subsection{Implementation Details}
\noindent\textbf{Diffusion Model.} We implement~\modelname based on the official code\footnote{https://github.com/openai/improved-diffusion} of improved diffusion~\cite{nichol2021improved}. $\theta$ is the default U-Net architecture with $128$ channels, $3$ ResNet blocks per layer, and the learn sigma flag disabled. For hyper-parameters of diffusion models, we use $4000$ diffusion steps with the cosine noise scheduler in all experiments, where the KL loss is not used.  

\noindent\textbf{Metric Function.} We use the LPIPS loss as a prototype metric function following consistency model~\cite{song2023consistency}, where we replace all MaxPooling layers of the LPIPS backbone with AveragePooling operations. 

\noindent\textbf{Training.} Training is conducted on $8$ V100 GPUs with $32$GB GPU RAM, where the batch size for each GPU is $16$, leading to $128$ accumulated batch size. We set learning rate to $1e^{-4}$ with no learning rate decay. When computing loss functions, $\lambda^\epsilon$, $\lambda^{x_0}$, and $\lambda^\mu$ are set to $1.0$ while $\lambda^{lpips}$ is set to $0.1$. We train the model for $400$k iterations and perform sampling and evaluation with the gap of $20$k and $100$k when the iteration is less than and higher than $100$k, respectively. For each model, we report the best result among all the evaluated checkpoints. 

\noindent\textbf{Sampling.} When sampling, we use the DDIM sampler and re-space the diffusion step to $100$. For each checkpoint, we sample $50$k images for CIFAR10 and $10$k images for other datasets and compute the evaluation metrics with respect to the training dataset.

\subsection{Experiment Settings}
\noindent\textbf{Datasets.} We conduct experiments on the CIFAR10, CelebAHQ, LSUN Bedroom, and ImageNet datasets. We train the model with image size $32\times32$ on CIFAR10 and $64\times64$ on the others.

\noindent\textbf{Metrics.} We compare models with Fr\'{e}chet Inception Distance (FID), sFID, and Inception Score (IS). FID and sFID evaluate the distributional similarity between the generated and training images. FID is based on the \texttt{pool\_3} feature of Inception V3~\cite{szegedy2016rethinking}, while sFID uses \texttt{mixed\_6/conv} feature maps. sFID is more sensitive to spatial variability~\cite{nash2021generating}. IS evaluates the quality of generated images without the need of reference images. 

\noindent\textbf{Baselines.} 
We compare~\modelname with the following baselines.
\begin{itemize}
    \item \textbf{DDPM ($\epsilon$ mode)}. We train DDPM by letting the U-Net predict the added noise $\epsilon$, which is then used to generate images with the DDIM sampler.
     \item \textbf{DDPM ($x_0$ mode)}. It is similar to DDPM ($\epsilon$ mode), where the U-Net predicts the clean image $x_0$.
     \item \textbf{DDPM ($x_0$) + LPIPS}. We add the LPIPS loss in the training of the DDPM ($x_0$ mode). 
     This model is to verify whether adding metric functions can improve the performance of DDPM.
     \item \textbf{Dual Diffusion}. We re-implement the Dual Diffusion Model based on the official code of the improved diffusion and then train the model with the same setting as other baselines.
     \item \textbf{Dual Diffusion + LPIPS}. We train the re-implemented Dual Diffusion Model by adding the LPIPS loss to its $x_0$ prediction. 
     This model is to verify whether adding metric functions can improve the performance of the Dual Diffusion Model.
\end{itemize}

We compare the proposed \modelname with the above baselines by conducting two experiments.
\begin{itemize}
    \item \textbf{\modelname}. We train \modelname with the same settings as other baselines. 
    This experiment is to demonstrate the performance of the vanilla \modelname without adding any metric function.
    \item \textbf{\modelname + LPIPS}. We add the LPIPS loss to the $x_0^\prime$ head of \modelname to verify whether the new diffusion model architecture enables the successful application of the LPIPS loss. 
\end{itemize}

\begin{figure*}[t]
    \centering
    \begin{subfigure}[b]{0.45\linewidth}
         \centering
         \includegraphics[width=\textwidth]{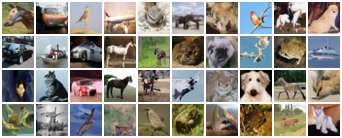}
         \caption{CIFAR10 samples. FID=$6.40$}
     \end{subfigure}
     \hfill
     \begin{subfigure}[b]{0.45\linewidth}
         \centering
         \includegraphics[width=\textwidth]{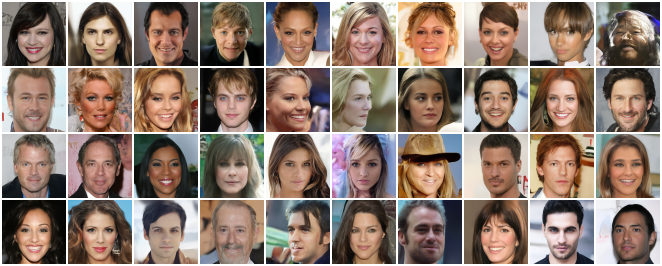}
         \caption{CelebAHQ samples. FID=$4.95$}
     \end{subfigure}
     \hfill
     \begin{subfigure}[b]{0.45\linewidth}
         \centering
         \includegraphics[width=\textwidth]{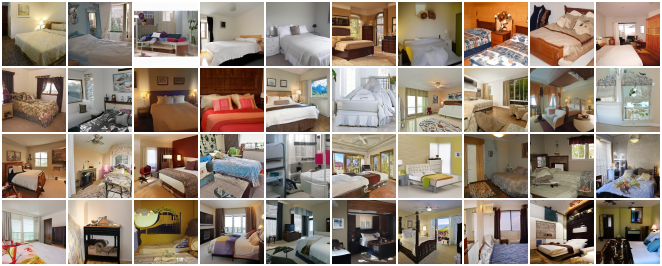}
         \caption{LSUN Bedroom samples. FID=$5.17$}
     \end{subfigure}
     \hfill
     \begin{subfigure}[b]{0.45\linewidth}
         \centering
         \includegraphics[width=\textwidth]{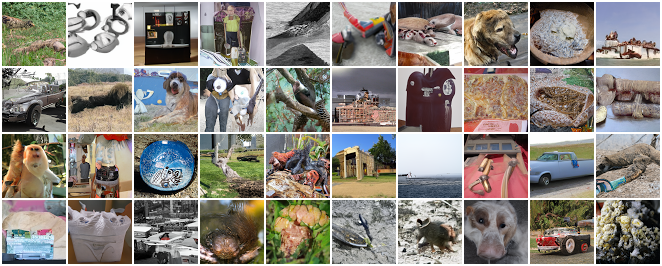}
         \caption{ImageNet samples. FID=$27.54$}
     \end{subfigure}
     \hfill
    \caption{Unconditional samples from \modelname trained with the LPIPS loss on the experimented datasets.
    }
    \label{fig:sample}
    \vspace{-3pt}
\end{figure*}

\begin{table}[t]
    \centering
    \setlength\tabcolsep{12pt}
    \resizebox{1\linewidth}{!}{
    \begin{tabular}{l|cc|cc}
    \toprule
    \multirow{2}*{\textbf{Model}} & \multicolumn{2}{c|}{\textbf{CIFAR10}} & \multicolumn{2}{c}{\textbf{CelebAHQ}} \\ 
    \cline{2-5} & FID$\downarrow$ & IS$\uparrow$ & FID$\downarrow$ & IS$\uparrow$ \\
    \midrule
    \modelname               & 6.80 & \textbf{8.88} & \underline{5.33} & \textbf{2.47} \\
    \modelname + LPIPS (VGG) & \textbf{6.40} & 8.69 & \textbf{4.95} & 2.37 \\

    \midrule
    \modelname + ResNet      & \underline{6.64} & \underline{8.75} & 5.67 & 2.41 \\ 
    \modelname + Inception   & 6.94 & \underline{8.75} & 5.52 & 2.41 \\
    \modelname + Swin        & 6.98 & 8.72 & 5.55 & 2.40 \\

    \bottomrule
    \end{tabular}
    }
    \caption{Performance comparison between pre-trained backbones of the metric functions.}
    \label{tab:ablation_arch}
\end{table}

\begin{table}[t]
    \centering
    \setlength\tabcolsep{6pt}
    \resizebox{1\linewidth}{!}{
    \begin{tabular}{l|cc|cc}
    \toprule
    \multirow{2}*{\textbf{Model}} & \multicolumn{2}{c|}{\textbf{CIFAR10}} & \multicolumn{2}{c}{\textbf{CelebAHQ}} \\ 
    \cline{2-5} & FID$\downarrow$ & IS$\uparrow$ & FID$\downarrow$ & IS$\uparrow$ \\
    \midrule
    \modelname               & 6.80 & \textbf{8.88} & 5.33 & \textbf{2.47} \\
    \modelname + LPIPS       & \underline{6.40} & 8.69 & \underline{4.95} & 2.37 \\

    \midrule
    \multicolumn{5}{l}{$\mathbf{\phi}$ \textbf{\textit{architectures}}} \\
    \midrule
    Fix-Res CNN              & \underline{6.40} & \underline{8.78} & 5.07 & \textbf{2.47} \\
    Fix-Res CNN + LPIPS      & \textbf{6.28} & 8.67 & \textbf{4.64} & \underline{2.44} \\
    \midrule
    \multicolumn{5}{l}{$\mathbf{\phi}$ \textbf{\textit{inputs}}} \\
    \midrule
    Fix-Res CNN + \texttt{cancat(}$x_0^\star$, $\epsilon^\prime$\texttt{)}   & 7.08 & 8.66 & 5.78 & 2.42 \\

    \bottomrule
    \end{tabular}
    }
    \caption{Performance comparison between different $\phi$ architectures and input settings.}
    \label{tab:ablation_phi}
\end{table}

\begin{table}[t]
    \centering
    \setlength\tabcolsep{7pt}
    \resizebox{1\linewidth}{!}{
    \begin{tabular}{l|cc|cc}
    \toprule
    \multirow{2}*{\textbf{Model}} & \multicolumn{2}{c|}{\textbf{CIFAR10}} & \multicolumn{2}{c}{\textbf{CelebAHQ}} \\ 
    \cline{2-5} & Step 10 & Step 100 & Step 10 & Step 100 \\
    \midrule
    DDPM ($\epsilon$ mode)   & 16.57 & \underline{6.79} & \underline{27.76} & 6.34 \\
    \modelname               & \underline{14.91} & 6.80 & 28.67 & \underline{5.32} \\
    \modelname + LPIPS       & \textbf{13.75} & \textbf{6.40} & \textbf{27.36} & \textbf{4.95} \\
    \bottomrule
    \end{tabular}
    }
    \caption{FID comparison between different sampling steps.}
    \label{tab:ablation_sample}
\end{table}

\subsection{Main Results}

\noindent\textbf{Overall Performance.} We compare the unconditional image generation performance of \modelname with baselines in Table~\ref{tab:sota}. We additional list the results in existing papers for reference, which are not comparable since they use different training and sampling settings. For CelebAHQ, LSUN Bedroom, and ImageNet, \modelname + LPIPS achieves the best FID and sFID scores among all the compared methods, where \modelname alone is the close next on CelebAHQ and LSUN Bedroom. This indicates that the architecture of \modelname is valuable compared with DDPM and Dual Diffusion Model since it produces better results than others on many datasets. More importantly, the improved performance achieves by \modelname + LPIPS indicates that adding metric functions such as LPIPS is a meaningful strategy to improve the performance of diffusion models, where \modelname shows a feasible diffusion model architecture design that can make it work. We notice that DDPM in $\epsilon$ mode achieves the highest IS score on LSUN Bedroom and ImageNet, while the FID and sFID scores are worse than \modelname + LPIPS. This may indicate that training with the LPIPS loss can improve the diversity of the generated images and make the overall distribution of the generated images closer to the training dataset, at a expense of slight degradation of the single image quality. 
A possible explanation is that the LPIPS loss enables the diffusion model to better capture and generate meaningful semantics because its VGG backbone is trained for recognition. The improved semantic correctness trades off a slight degradation of pixel-level image quality with the better data distribution alignment.
Fig.~\ref{fig:sample} shows the generated images by \modelname assisted with the LPIPS loss.

\noindent\textbf{Metric Function Effectiveness.} Table~\ref{tab:lpips} compares the performance of diffusion models with and without the LPIPS loss. We list the FID increase or drop after the usage of the LPIPS loss on DDPM in $x_0$ mode, Dual Diffusion Model, and \modelname. DDPM ($x_0$ model) and Dual Diffusion Model achieve improved performance (reduced FID score) after applying the LPIPS loss on CIFAR10 and ImageNet. However, the results are inconsistent on the other two datasets. The proposed \modelname achieves consistently improved performance among all the compared datasets. This indicates that the architectural design of \modelname enables the effective application of the LPIPS loss on diffusion model training.



\subsection{Ablation Study}
We conduct an ablation study on metric function backbones, the architecture as well as the input settings of network $\phi$ in \modelname, and the DDIM sampling steps.

\noindent\textbf{Metric Function Backbones.} Our experimental results have demonstrated that the LPIPS loss can improve the performance of diffusion models. Table~\ref{tab:ablation_arch} compares the LPIPS loss with other metric function backbones. We use the ResNet~\cite{he2016deep}, Inception v3~\cite{szegedy2016rethinking}, and Swin Transformer~\cite{liu2021swin} pre-trained on the ImageNet dataset. Similar to the LPIPS loss, we first extract their features on images at different layers, which is then used to compute the mean square error between corresponding feature maps. We find that ResNet can improve the performance of CIFAR10 generation while all the other pre-trained networks do not work. We conjecture that the VGG network~\cite{simonyan2014very} used by the LPIPS loss does not use residual connections, which may make the extracted features contain more semantic information. Similar observations have also been made by~\citet{karras2020analyzing}. We leave the discovery of more powerful metric functions and other strategies for improving the diffusion model training as future work.

\noindent\textbf{$\phi$ Architectures and Inputs.} \modelname uses the U-Net with the same architecture in $\theta$ and $\phi$. Although U-Net works well, Table~\ref{tab:ablation_phi} shows the results with a different architecture of $\phi$. We remove all the downsampling and the upsampling layers of the U-Net, resulting in a network with the fixed resolution in all layers, denoted as Fix-Res CNN. We remove a few attention layers in the Fix-Res CNN to make it easier for training. The Fix-Res CNN is a network with fewer network parameters but involves more FLOPs compared with the original U-Net used by $\phi$. We find that this network achieves better results with and without the LPIPS loss than the original U-Net in terms of FID, which indicates a direction to improve \modelname. With the motivation that $\epsilon^\prime$ can influence the appearance of $x_0^\star$ according to Eq.~\ref{eq:5}, we also attempt to take the concatenation of $x_0^\star$ and $\epsilon^\prime$ as the input to train $\phi$, which we find does not work well in terms of FID but slightly improves the IS score.

\noindent\textbf{Sampling Steps.} Table~\ref{tab:ablation_sample} compares the FID of DDPM in the $\epsilon$ mode and \modelname with/without the LPIPS loss on sampling steps $10$ and $100$. 
\modelname works equally well on small and large sampling steps in terms of enabling the successful application of the LPIPS loss. Note that on different datasets,  a comparably larger performance gain may be achieved by either smaller or larger sampling steps.

%% file: sections/conclusion.tex
\section{Conclusion}
\label{sec:conclusion}

In this paper, we study how to use metric functions to improve the performance of image diffusion models. To this end, we propose \modelname, which uses two cascaded networks to predict the added noise and the clean image, respectively. This architecture design addresses the issue of the dual diffusion model where the LPIPS loss affects the noise prediction.
Experimental results on several datasets show that \modelname assisted with the LPIPS loss achieves the state-of-the-art results.